\PassOptionsToPackage{numbers}{natbib}
\documentclass{article}

\usepackage[preprint]{arxiv}

\usepackage[utf8]{inputenc} 
\usepackage[T1]{fontenc}    
\usepackage{hyperref}       
\usepackage{url}            
\usepackage{booktabs}       
\usepackage{amsfonts}       
\usepackage{nicefrac}       
\usepackage{microtype}      
\usepackage{xcolor}         
\usepackage{amsmath}

\usepackage{doi}
\usepackage{graphicx}
\usepackage{subcaption}
\usepackage{wrapfig}
\usepackage[table]{xcolor}
\usepackage{caption}
\usepackage{microtype}

\usepackage{hyperref}

\definecolor{LighterGray}{gray}{0.95}
\definecolor{LightGreen}{RGB}{232,244,234}
\definecolor{GainColor}{HTML}{228B22} 

\usepackage[labelfont=bf]{caption} 

\newcommand{\bparagraph}[1]{\vspace{2pt}\noindent\textbf{#1}}

\title{STRIVE: Structured Spatiotemporal Exploration for Reinforcement Learning in Video Question Answering}

\author{
    \textbf{Emad Bahrami}\textsuperscript{1,4}\thanks{Work was done during an internship at Microsoft.} \quad
    \textbf{Olga Zatsarynna}\textsuperscript{1,4} \quad
    \textbf{Parth Pathak}\textsuperscript{3} \\ \vspace{0.15cm}
    \textbf{Sunando Sengupta}\textsuperscript{2} \quad
    \textbf{Juergen Gall}\textsuperscript{1,4} \quad
    \textbf{Mohsen Fayyaz}\textsuperscript{2} \\ \vspace{0.3cm}
    \textnormal{\textsuperscript{1}University of Bonn \quad \textsuperscript{2}Microsoft \quad \textsuperscript{3}Meta} \\[ -7pt]
    \textnormal{\textsuperscript{4}Lamarr Institute for Machine Learning and Artificial Intelligence}
}

\begin{document}

\maketitle

\begin{abstract}

We introduce STRIVE (SpatioTemporal Reinforcement with Importance-aware Variant Exploration), a structured reinforcement learning framework for video question answering. While group-based policy optimization methods have shown promise in large multimodal models, they often suffer from low reward variance when responses exhibit similar correctness, leading to weak or unstable advantage estimates. STRIVE addresses this limitation by constructing multiple spatiotemporal variants of each input video and performing joint normalization across both textual generations and visual variants. By expanding group comparisons beyond linguistic diversity to structured visual perturbations, STRIVE enriches reward signals and promotes more stable and informative policy updates.
To ensure exploration remains semantically grounded, we introduce an importance-aware sampling mechanism that prioritizes frames most relevant to the input question while preserving temporal coverage. This design encourages robust reasoning across complementary visual perspectives rather than overfitting to a single spatiotemporal configuration.
Experiments on six challenging video reasoning benchmarks—including VideoMME, TempCompass, VideoMMMU, MMVU, VSI-Bench, and PerceptionTest—demonstrate consistent improvements over strong reinforcement learning baselines across multiple large multimodal models. Our results highlight the role of structured spatiotemporal exploration as a principled mechanism for stabilizing multimodal reinforcement learning and improving video reasoning performance.

\end{abstract}

\section{Introduction}
\label{sec:intro}

Recent advances in large multimodal models (LMMs) have significantly improved performance on video question answering (VQA)~\cite{li2024llavaonevision, bai2025qwen2, cheng2024videollama2}, where models must jointly reason over visual and textual inputs to answer complex queries about dynamic scenes. While supervised fine-tuning has driven much of this progress, its reliance on large-scale annotated data limits scalability and often fails to fully exploit the structured nature of video inputs. As a result, post-training reinforcement learning (RL) has emerged as a promising alternative, enabling models to refine their reasoning behavior through task-driven reward signals without requiring additional human annotation. Group-based policy optimization methods have recently shown strong empirical gains in both language and multimodal domains~\cite{shao2024deepseekmath, deepvideo2025r1}. By sampling multiple candidate responses for a given input and computing relative advantages within each group, these approaches encourage policies to favor higher-quality outputs. However, when applied to video-based reasoning, existing formulations typically treat the video input as fixed and only diversify textual generations. 
This overlooks a fundamental property of video: its rich spatiotemporal structure offers multiple valid perspectives from which reasoning can emerge. A key challenge arises when the reward variance within a sampled group is low—for example, when most generated responses are uniformly correct or incorrect. In such cases, the resulting advantage estimates become weak or unstable, limiting effective policy improvement. While prior work has addressed this issue through modifications of normalization or reward shaping~\cite{liu2025drgrpo}, these solutions operate solely on textual outputs and do not leverage the visual modality itself as a source of structured diversity.

\begin{figure}[t]
    \centering
    \includegraphics[width=0.65\columnwidth, trim={2cm 0cm 2cm 1cm}]{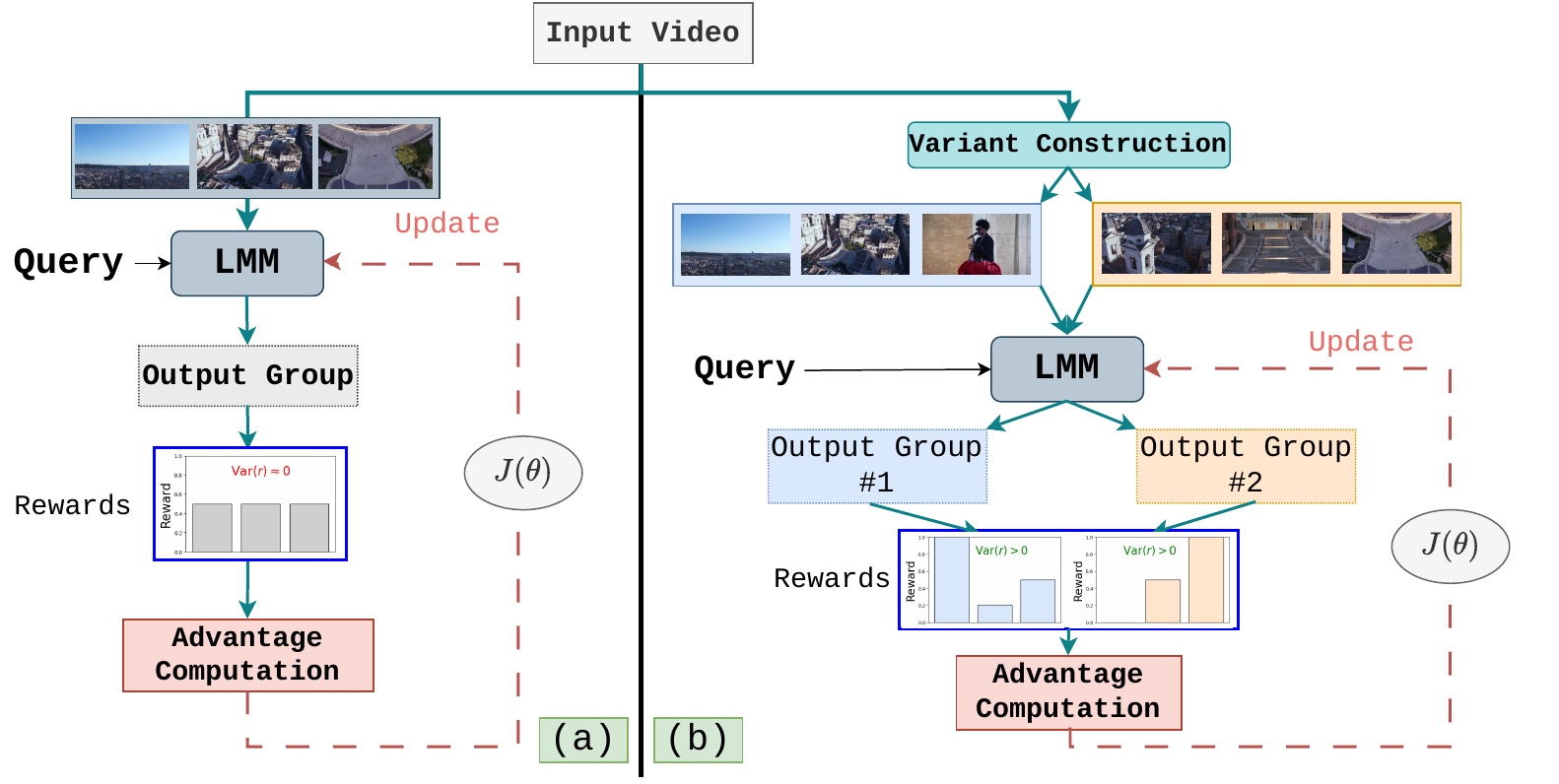}
    \caption{\small \textbf{STRIVE prevents advantage collapse in video reinforcement learning.} (a) Standard policy optimization samples multiple textual responses from a single, fixed video representation. When these generated responses lack diversity, the reward variance can drop to zero ($\text{Var}(r) = 0$), stalling the gradient of the objective $J(\theta)$. (b) Our proposed STRIVE framework employs Variant Construction to systematically generate multiple spatiotemporal visual variants of the input video. By computing joint advantages across both textual generations and varied visual perspectives, STRIVE empirically maintains a consistently positive reward variance ($\text{Var}(r) > 0$), delivering informative updates to the policy $J(\theta)$.}
    \label{fig:teaser}
\end{figure}

In this work, we introduce \textbf{STRIVE (SpatioTemporal Reinforcement with Importance-aware Variant Exploration)}, a structured reinforcement learning framework designed to exploit the inherent spatiotemporal structure of video. As illustrated in Fig.~\ref{fig:teaser}, instead of grouping only textual generations, STRIVE constructs joint groups across both textual responses and multiple spatiotemporal variants of the same input video. By generating complementary visual perspectives and organizing them into a dual-grouping mechanism, our method goes beyond simply increasing reward variance. These structured input perturbations act as an implicit consistency regularizer, encouraging the policy to maintain cross-variant agreement and promoting stronger reliance on visual evidence.
To ensure that visual diversity remains semantically meaningful, we further introduce a temperature-scaled, \textbf{importance-aware sampling strategy}. Rather than treating all frames equally, STRIVE estimates per-frame relevance scores conditioned on the input question using cross-modal embeddings. Frames with higher semantic alignment to the query are prioritized during variant construction, enabling targeted exploration of question-critical events. To prevent overfitting to isolated moments and preserve global narrative coherence, we incorporate a segmented sampling scheme that provides global temporal coverage while maintaining semantic focus. This design encourages the model to reason over both salient cues and broader contextual structure. 

Together, structured spatiotemporal grouping and importance-aware exploration provide a principled mechanism for stabilizing reinforcement learning in video multimodal models. Experiments across six challenging benchmarks—including VideoMME~\cite{fu2025videomme}, TempCompass~\cite{liu2024tempcompass}, VideoMMMU~\cite{hu2025videommmu}, MMVU~\cite{zhao2025mmvu}, VSI-Bench~\cite{yang2025vsibench}, and PerceptionTest~\cite{patraucean2023perception}—demonstrate consistent improvements over strong reinforcement learning baselines. Our results highlight the importance of explicitly leveraging spatiotemporal structure when performing policy optimization for video reasoning tasks.

\section{Related Work}
\label{sec:related_work}

\subsection{Large Multimodal Models (LMMs)}
Large multimodal models (LMMs) have rapidly advanced in recent years, extending large language models with visual encoders to handle complex reasoning tasks over images and videos. Early progress was primarily driven by image-based tasks and visual instruction tuning~\cite{liu2024improved, tong2024cambrian, liu2023visual, li2024llavaonevision}, which improved OCR, grounding, and reasoning. More recently, research has shifted toward video understanding. Models such as Video-LLaMA~\cite{zhang2023videollama}, Video-LLaVA~\cite{lin2023videollava}, and Video-ChatGPT~\cite{maaz2024videochatgpt} successfully extend instruction-tuned LMMs to temporal reasoning. This shift has led to significant improvements across fine-grained spatiotemporal benchmarks, including Video-MME~\cite{fu2025videomme}, Video-MMMU~\cite{hu2025videommmu}, TempCompass~\cite{liu2024tempcompass}, VSI-Bench~\cite{yang2025vsibench}, and PerceptionTest~\cite{patraucean2023perception}. Despite this strong progress, open models still lag behind proprietary models such as GPT-4o~\cite{hurst2024gpt4o} and Gemini 1.5~\cite{team2024gemini}, whose detailed training methods and architectures remain undisclosed.

\subsection{Post-Training for LMMs}
Reinforcement learning (RL) has become the standard post-training paradigm for aligning large language models (LLMs) with human preferences~\cite{stiennon2020learning,ouyang2022training,rafailov2023direct}. Methods such as PPO~\cite{schulman2017proximal}, DPO~\cite{rafailov2023direct}, and GRPO~\cite{shao2024deepseekmath} have significantly enhanced reasoning capabilities. Building on this success, RL is increasingly being applied to large multimodal models (LMMs). In particular, GRPO has emerged as a favored approach for multimodal alignment, as it enables memory-efficient, group-wise optimization over multiple responses.

Recent efforts have specifically adapted GRPO to tackle the unique challenges of video understanding and temporal reasoning. For instance, ArrowRL~\cite{xue2025arrowrl} introduces a ``reverse reward'' mechanism that encourages models to distinguish forward from reversed video sequences, instilling an Arrow-of-Time awareness. Similarly, Video-R1~\cite{feng2025videor1} proposes T-GRPO, which contrasts reasoning performance on ordered versus shuffled frames to enhance temporal perception. Extending this temporal focus, Time-R1~\cite{wang2025timer1} formulates an RL-based framework specifically for temporal video grounding. To stabilize training in these complex temporal settings, DeepVideo-R1~\cite{deepvideo2025r1} introduces Reg-GRPO, a regression-style variant paired with difficulty-aware video augmentations to mitigate vanishing advantages. A closely related emerging trend utilizes visual noise injection to enhance policy exploration. For example, NoisyRollout~\cite{liu2025noisyrollout} mixes training trajectories from clean and distorted images using a noise annealing schedule to inject perceptual diversity without altering the RL objective. Similarly, NoisyGRPO~\cite{qiu2025noisygrpo} perturbs visual inputs with Gaussian noise and employs a Bayesian advantage estimation framework to explicitly encourage LMMs to favor visually grounded reasoning over noisy trajectories. Beyond temporal alignment and noise injection, complementary works have explored advanced reward shaping and explicit reasoning strategies to improve overall video comprehension. On the algorithmic side, R1-ShareVL~\cite{yao2025sharevl} expands each query into variants and hierarchically shares advantages, while VideoChat-R1~\cite{li2025videochatr1} leverages multi-task reinforcement fine-tuning for joint spatiotemporal perception. Concurrently, models are being trained to actively filter visual information; Chain-of-Frames~\cite{ghazanfari2025chainofframes} leverages step-by-step reasoning to dynamically select relevant frames, and VideoAuto-R1~\cite{liu2026videoautor1} employs an adaptive strategy to autonomously determine when such explicit reasoning is required before generating an answer.

While these methods advance multimodal alignment, they largely retain the standard GRPO formulation and address temporal perception primarily through reward engineering or auxiliary augmentations. ArrowRL and Video-R1 rely on handcrafted contrastive rewards based on forward/reverse or ordered/shuffled variants, while DeepVideo-R1, NoisyRollout, and NoisyGRPO dynamically adjust the difficulty of the visual inputs or rely on unstructured noise augmentations outside of the grouping mechanism. In contrast, our proposed framework, STRIVE, fundamentally restructures the advantage calculation itself. Rather than abandoning standard relative optimization for a regression loss or relying on unstructured noise injection, STRIVE integrates \emph{spatiotemporal video variants} directly into a joint grouping strategy. By forming groups that combine textual responses with a question-aware, importance-based selection of frames, STRIVE enriches reward variance. This structured grouping strategy provides a reliable learning signal in low-variance regimes, leading to consistent gains over prior GRPO-based post-training approaches on video question answering benchmarks.

\section{Method}
\label{sec:method}

\begin{figure}[t]
    \centering
    \includegraphics[width=\linewidth]{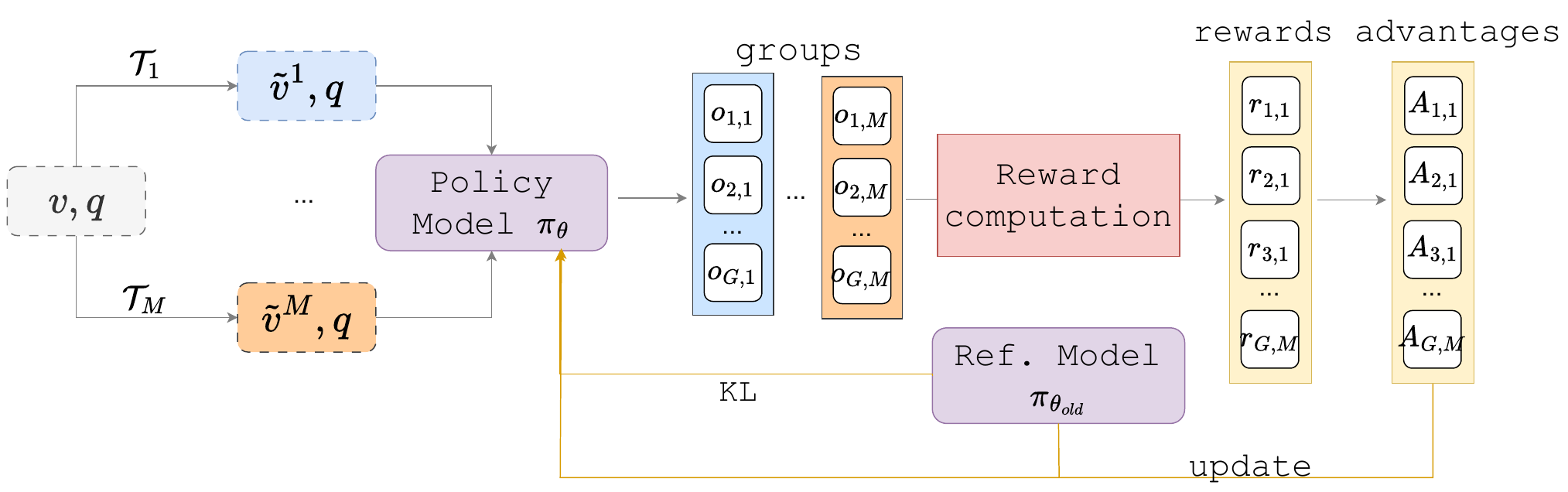}
    \caption{\textbf{Overview of our STRIVE Framework}. Given a video input $v$ and question $q$, a set of transformations $\mathcal{T}_i$ is applied to generate video variants $\tilde{v}_i$. Each variant is paired with the question to form input tuples $(\tilde{v}_i, q)$. The policy model produces $G$ textual responses for each of the $M$ variants, resulting in a $G \times M$ matrix of responses $o_{i,j}$. These are evaluated by a reward model to obtain $r_{i,j}$. Advantages are then jointly normalized across the entire $G \times M$ pool, enriching reward variance.}
    \label{fig:strive_overview}
\end{figure}

\subsection{Preliminaries: GRPO and Advantage Collapse}
\label{sec:preliminaries}

Group Relative Policy Optimization (GRPO)~\cite{shao2024deepseekmath} is an efficient variant of Proximal Policy Optimization (PPO)~\cite{schulman2017proximal} that eliminates the need for a separate value function network. Instead, it estimates advantages directly from the relative rewards of multiple sampled outputs. Given a question $q$ and a video input $v$, the model samples multiple candidate responses $\{o_i\}_{i=1}^G$ from the policy $\pi_{\theta}$. Each output receives a reward $r_i$, and the policy is optimized using the objective:

\begin{equation}
\begin{aligned}
J_{\text{GRPO}}(\theta)  = \frac{1}{G} \sum_{i=1}^{G} \Bigg[ 
\min \!\Bigg( 
\frac{\pi_{\theta}(o_i \mid q, v)}{\pi_{\theta_{\text{old}}}(o_i \mid q, v)} A_i, \;
& \text{clip}\!\Big( 
\frac{\pi_{\theta}(o_i \mid q, v)}{\pi_{\theta_{\text{old}}}(o_i \mid q, v)}, \; 1-\epsilon, \; 1+\epsilon 
\Big) A_i 
 \Bigg) \Bigg] \\ 
& - \beta \, \mathbb{D}_{\text{KL}}\!\big[ \pi_{\theta} \;||\; \pi_{\text{ref}} \big]        
\end{aligned}
\label{eq:grpo_obj}
\end{equation}

where $A_i$ denotes the normalized advantage, computed over the group:
\begin{equation}
A_i = \frac{r_i - \text{mean}(r_1, \dots, r_G)}{\text{std}(r_1, \dots, r_G)}.   
\label{eq:grpo_advantage}
\end{equation}

\bparagraph{The Zero-Advantage Bottleneck.} 
While GRPO effectively encourages relatively better responses, it exhibits a critical bottleneck when the intra-group reward variance approaches zero (e.g., when responses to a video QA prompt are uniformly correct or incorrect). When variance approaches zero, advantages either explode (due to division by a small denominator) or collapse to zero when rewards are identical. To mitigate this, Dr.GRPO~\cite{liu2025drgrpo} removes the standard deviation term entirely. However, this introduces a secondary bias: rare correct responses in an otherwise incorrect group are severely down-weighted. Increasing the intrinsic reward variance resolves both issues simultaneously, preventing exaggerated updates while ensuring high-quality responses stand out. Beyond variance amplification, we argue that the core limitation of text-only grouping lies in its inability to expose the model to semantically diverse visual evidence within a group. When all samples share identical visual inputs, group-based normalization only compares linguistic variability. By introducing structured visual perturbations, we expand the comparison space beyond mere textual differences. This allows the RL algorithm to reward models that genuinely anchor their text responses in diverse visual evidence, enforcing a deeper consistency between what the model sees and what it generates.

\subsection{The STRIVE Framework: A Dual-Grouping Paradigm}
\label{sec:strive_framework}

Traditional group-based RL methods rely exclusively on \textbf{output diversity}---generating multiple textual responses for a fixed input. We propose a fundamental conceptual shift for multimodal tasks: moving diversity to the input space. The visual modality offers an inherent source of structured, spatiotemporal diversity. We introduce \textbf{STRIVE (SpatioTemporal Reinforcement with Importance-aware Variant Exploration)}, which restructures the advantage calculation by constructing a joint optimization space over both inputs and outputs.

As illustrated in Fig.~\ref{fig:strive_overview}, given a video $v$ and question $q$, STRIVE applies $M$ transformations to generate diverse video variants $\{\tilde{v}_i\}_{i=1}^M$. Crucially, for \emph{each} variant, the policy generates $G$ textual responses. This transforms the standard 1D array of responses into a $G \times M$ matrix $\{o_{i,j}\}_{i=1, j=1}^{G,M}$ with corresponding rewards $\{r_{i,j}\}$. STRIVE optimizes the standard GRPO objective (Eq.~\ref{eq:grpo_obj}) over this flattened dual-group, defining the joint advantage $A_{i,j}$ as:
\begin{equation}
A_{i,j} = \frac{r_{i,j} - \text{mean}(\{r_{i,j}\}_{i=1, j=1}^{G,M})}{\text{std}(\{r_{i,j}\}_{i=1, j=1}^{G,M})}.   
\label{eq:strive_advantage}
\end{equation}

\subsection{Importance-Aware Variant Construction}
\label{sec:transformations}

To construct the visual variants and rigorously evaluate our framework, we explore three distinct categories of spatiotemporal transformations. First, \textbf{Deterministic Temporal} transformations select frames using standard strided sampling with staggered temporal offsets to create non-overlapping visual subsets without any guided semantic focus. Second, \textbf{Stochastic} transformations inject visual diversity blindly through spatial augmentations, including color jittering (adjusting brightness, contrast, and saturation) and minor affine transformations (slight rotations, translations, and scaling). Finally, our primary method, \textbf{Importance-Based Grouping}, guides exploration by prioritizing frames that are semantically relevant to the user's query.

\bparagraph{Importance-Based Scoring.} 
To construct semantically-aware groups, we treat the video as a sequence of frames $v = \{v_1, v_2, \dots, v_T\}$. We first compute a temporal relevance score $s_t = \text{rel}(v_t, q)$ for each frame $v_t$ given the query $q$. We evaluate two distinct instantiations for the relevance function. For \textbf{Cross-Modal Similarity (Sim)}, we leverage the inherently aligned representation space of the underlying multimodal model. The score $s_t = \hat{z}_t^\top \hat{z}_q$ is defined as the cosine similarity between the $L_2$-normalized pooled visual embeddings ($\hat{z}_t$) and the averaged textual token embeddings ($\hat{z}_q$) extracted directly from the network. This efficiently quantifies how well a frame visually aligns with the semantic context of the question. Alternatively, using \textbf{Gradient-Based Attribution (Grad)}, we derive temporal saliency directly from the policy model. We compute the gradient of the language modeling loss—restricted strictly to the tokens corresponding to the user's question—with respect to the input video pixels. These spatiotemporal gradients are then aggregated to identify the most critical visual features.

\bparagraph{Temperature-Scaled Sampling.}
Using the computed scores $\{s_t\}_{t=1}^T$, we partition the video into $K$ equal temporal segments, where $K$ is set to our fixed sampling budget. This segmentation ensures global temporal coverage, preventing the model from overfitting to isolated events and preserving the broader narrative context. In policy optimization, maintaining exploration is essential for preventing premature convergence~\cite{sutton2018reinforcement}. If STRIVE were to deterministically select the single highest-scoring frame from each segment (pure exploitation), it would severely restrict visual diversity, inadvertently recreating the low-variance bottleneck. To enforce structured exploration, we employ \textbf{temperature-scaled softmax sampling}~\cite{sutton2018reinforcement}. For a segment spanning frames from index $t=s_i$ to $e_i$, we sample the representative frame index $t_i$ stochastically using:
\begin{equation}
P(t) = \frac{\exp(s_t / \tau)}{\sum_{u=s_i}^{e_i} \exp(s_u / \tau)}, \quad t_i \sim \text{Categorical}(P).
\label{eq:weighted_sampling}
\end{equation}
The temperature $\tau > 0$ controls the exploration-exploitation trade-off. A lower $\tau$ sharpens the distribution to exploit the most query-aligned frames, while a higher $\tau$ encourages broader temporal exploration. This probabilistic approach empirically ensures that the model predominantly attends to relevant visual evidence while preserving the intra-group diversity necessary for meaningful relative advantage estimates.

\section{Experiments}
\label{sec:experiments}

\subsection{Experimental Setup}

\bparagraph{Implementation Details.} 
We employ Qwen2.5-VL-7B and Qwen3-VL-8B as the base Large Multimodal Models (LMMs) for our reinforcement learning experiments. During training, we extract a maximum budget of 64 frames per video. From this budget, our STRIVE variant construction module samples exactly 16 frames per forward pass to compute visual features. To ensure a strictly fair comparison, the standard GRPO baseline also extracts exactly 16 frames per forward pass, utilizing standard uniform temporal sampling. All models are trained for 1,000 RL steps using a subset of the Video-R1 dataset~\cite{feng2025videor1}. More details are provided in the supplementary.

\begin{figure}[t!]
    \centering
    \includegraphics[width=0.9\columnwidth]{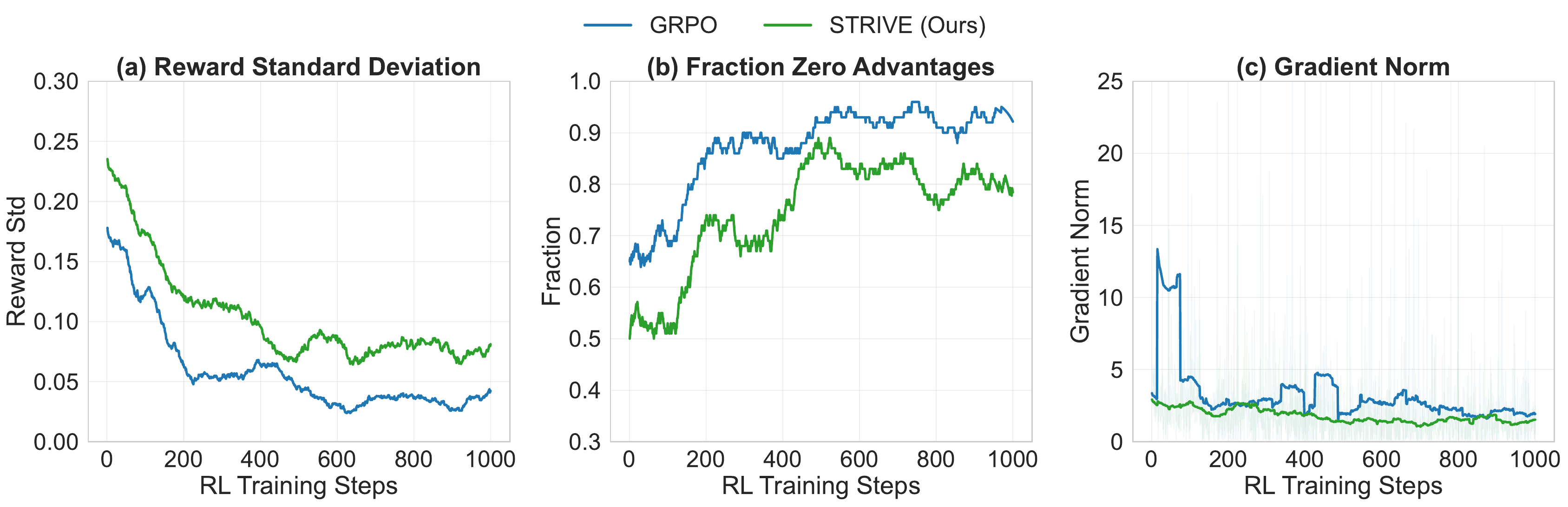}
    \caption{\textbf{Empirical analysis of optimization dynamics.} \textbf{(a)} Standard deviation of rewards, demonstrating that STRIVE maintains consistently higher variance than the GRPO baseline. \textbf{(b)} Fraction of zero-advantage updates, illustrating STRIVE's ability to prevent gradient starvation. \textbf{(c)} Gradient norm over training steps. 
    }
    \label{fig:analysis_plot}
\end{figure}

\bparagraph{Empirical Analysis of Advantage Collapse}
As described in Sec.~\ref{sec:preliminaries}, standard GRPO suffers from a critical bottleneck when reward variance within a group collapses, leading to uninformative advantages and unstable optimization. To empirically validate that our dual-grouping paradigm resolves this issue, we analyze the training dynamics of the baseline GRPO and our proposed STRIVE framework (Fig.~\ref{fig:analysis_plot}).

First, Fig.~\ref{fig:analysis_plot}a tracks the standard deviation of rewards over the RL training steps. While GRPO exhibits a steady decline in variance as textual responses within a group converge (i.e., becoming uniformly correct or incorrect), STRIVE maintains a consistently higher and healthier variance by leveraging the structured diversity of spatiotemporal visual variants.

This sustained variance directly prevents advantage collapse. As shown in Fig.~\ref{fig:analysis_plot}b, standard GRPO frequently encounters updates where the group reward standard deviation is exactly zero, resulting in a high fraction of zero-advantage steps that starve the policy of a learning signal. STRIVE effectively suppresses this fraction, ensuring a continuous and discriminative gradient.

Finally, the severe reduction in reward variance under standard GRPO actively destabilizes the optimization process. Because the standard deviation acts as the denominator in the advantage computation (Eq.~\ref{eq:grpo_advantage}), a vanishing variance artificially inflates the advantages. Fig.~\ref{fig:analysis_plot}c illustrates this phenomenon: GRPO exhibits spikes in the gradient norm, which are clearly visible in the volatile raw trajectory. In contrast, by systematically injecting structured spatiotemporal diversity into the input space, STRIVE naturally preserves a healthy variance bound. This ensures the advantage denominator remains strictly positive, yielding a significantly smoother and more stable gradient trajectory.

\begin{table}[t!]
    \centering
\caption{\textbf{Performance on six video understanding benchmarks.} To ensure a fair comparison, models are evaluated with a maximum of 32 frames. Our proposed STRIVE framework outperforms the standard GRPO baseline across both Qwen2.5-VL-7B and Qwen3-VL-8B architectures. Results marked with * are evaluated using publicly available model weights.}
\vspace{0.1cm}
    \label{tab:main_results}
    \resizebox{1.0\columnwidth}{!}{%
    \begin{tabular}{l|c|c|c|c|c|c|c}
        \toprule
        Model                                      & VideoMME (wo sub) & VSI-Bench & PerceptionTest (mc) & TempCompass & VideoMMMU & MMVU (mc) & avg \\ 
        \toprule
        GPT-4o~\cite{achiam2023gpt4o}              & 71.9              & 34.0         &           -       & 75.1        & 61.2      & 75.4      & - \\
        \midrule
        Gemini-1.5-Pro\cite{team2024gemini}        & 75.0              & 48.8         &       -           & 70.5        & -      & -      & - \\
        \midrule
        \midrule
        VILA-1.5-8B~\cite{lin2024vila}             & -              & 28.9         &   -    & -        & 20.8      & -      & - \\
        \midrule               
        LongVA-7B~\cite{zhang2024longva}           & 52.6              & 29.2      &   -    & -        & 23.9      & -      & - \\
        \midrule
        LLaVA-OV-7B~\cite{li2024llavaonevision}             & 58.2              & 32.4      &    57.1   & 69.5        & -      & 64.7      & - \\
        \midrule
        \midrule
        Qwen2.5-VL-7B-SFT~\cite{feng2025videor1}   & 55.4            & 33.3        & 68.4 &    69.9   & 49.4      & 63.5      & 56.6 \\
        \midrule        
        Arrow RL*~\cite{xue2025arrowrl}             &   60.9           &   36.1       & 66.8 &    72.5     &   47.9    &   62.4    & 57.7 \\  
        \midrule
        Video-R1~\cite{feng2025videor1}           &  59.3             & 35.8    &  67.8 &  73.2       &    52.3   &   63.8    &  58.6 \\   
        \midrule
        \midrule
        Qwen2.5-VL-7B                          & 57.8           & 34.3         &      68.2         & 72.8        & 48.6      & 63.8      & 57.6 \\
        \midrule
        GRPO                              & 61.1            & 35.9  &  67.7   & 75.1        & 49.5      & 64.6      & 59.0 \\ 
        \midrule
        \rowcolor{LightGreen}
        STRIVE (ours)                                      & 61.7              & 36.6  &  69.4    & 75.1        & 50.1      & 65.7      & \textbf{59.8}    \\ 
        \midrule
        \midrule
        Qwen3-VL-8B                          & 62.1           & 53.3         &  68.7    & 77.8        & 56.5      & 67.0      & 64.2 \\
        \midrule
        GRPO                              & 60.3            & 53.0         &  70.0    & 78.1        & 55.7      & 68.6      & 64.3 \\ 
        \midrule
        \rowcolor{LightGreen}
        STRIVE (ours)                          & 62.1            & 54.4         &  70.2    & 78.3        & 56.5      & 69.6      & \textbf{65.2}    \\ 
        
        \bottomrule
    \end{tabular}
    } 
\end{table}

\bparagraph{Main Results.} 
Table~\ref{tab:main_results} presents the main results of our method compared to LMM baselines and contemporary reinforcement learning approaches. We report average performance across six challenging benchmarks that collectively evaluate temporal reasoning, spatial understanding, and multimodal perception: VideoMME~\cite{fu2025videomme}, VSI-Bench~\cite{yang2025vsibench}, PerceptionTest~\cite{patraucean2023perception}, TempCompass~\cite{liu2024tempcompass}, VideoMMMU~\cite{hu2025videommmu}, and MMVU~\cite{zhao2025mmvu}. 

We first consider supervised fine-tuning (Qwen2.5-VL-7B-SFT) and existing reinforcement learning baselines (Arrow RL, Video-R1). Among these, Video-R1 achieves the strongest overall performance with an average score of 58.6, demonstrating the benefit of reinforcement learning over pure supervised training.

Next, we evaluate our reinforcement learning models built on the Qwen2.5-VL-7B architecture. Standard GRPO improves upon the base model, raising the average from 57.6 to 59.0, confirming the effectiveness of GRPO in aligning model outputs with video QA tasks. Our proposed STRIVE framework further boosts performance, achieving a new best average score of 59.8. These comprehensive improvements highlight the benefit of constructing spatiotemporal groups, which increase reward variance and encourage more informative exploration during reinforcement learning.

Furthermore, we evaluate the scalability of our approach using the more capable Qwen3-VL-8B architecture. While standard GRPO achieves a marginal increase in the overall average score (from 64.2 to 64.3), it results in performance drops on specific benchmarks, such as VideoMME (62.1 to 60.3) and VideoMMMU (56.5 to 55.7), compared to the base model. In contrast, STRIVE avoids these performance drops, maintaining or improving upon the base model's strong reasoning capabilities across all evaluated benchmarks. As a result, STRIVE achieves an overall average score of 65.2, outperforming the GRPO baseline by nearly a full point and confirming that our spatiotemporal grouping strategy scales effectively to stronger vision-language models.

\subsection{Ablation Study} 
To thoroughly analyze our framework, we conduct 500 steps of RL training for all ablation experiments. Table~\ref{tab:combined_ablations} summarizes the results across three key dimensions: variant construction strategies, scoring architectures, and the trade-off between visual and textual samples.

\bparagraph{Variant Construction Strategies.}
We systematically evaluate the three broad transformation categories introduced in our methodology (Table~\ref{tab:combined_ablations}a). Compared to the GRPO baseline, applying a deterministic temporal strategy yields mixed results and slightly lowers the average performance (56.8). Because this approach lacks stochasticity, it restricts the policy's exploration space, making the model prone to converging on sub-optimal behaviors. In contrast, introducing stochastic transformations acts as a catalyst for exploration by randomly diversifying the visual inputs, achieving notable gains on VSI-Bench and raising the overall average to 57.9. However, pure stochasticity explores blindly. In video QA, where critical visual evidence is often sparsely distributed, unguided random cropping risks discarding the exact frames needed to answer the query, potentially introducing misaligned reward signals. Our proposed importance-based grouping resolves this by providing \textit{guided} exploration. By prioritizing query-relevant frames, it ensures the model explores diverse visual perspectives without losing critical semantic context, achieving the highest overall average (58.1). Finally, we observe a distinct trade-off when complementing this temporal selection with spatial augmentations. While spatial perturbations introduce minor visual noise that slightly degrades performance on fine-grained semantic tasks like VideoMME and MMVU (resulting in a nominal drop in the overall average to 58.0), they act as a strong spatial regularizer, driving VSI-Bench to a high of 35.2.

\bparagraph{Impact of the Scoring Mechanism.}
In our importance-based grouping strategy, the mechanism $\mathcal{S}$ determines how semantic relevance is quantified (Table~\ref{tab:combined_ablations}b). We first compare gradient-based attribution against cross-modal embedding similarity. While deriving temporal saliency directly via gradients achieves a performance gain on VideoMME (61.1), it struggles to generalize to spatial tasks like VSI-Bench, resulting in a lower overall average (57.7). Consequently, we focus on cross-modal similarity. Using the policy model itself (e.g., Qwen2.5-VL-7B) to compute similarity yields the highest overall average (58.6), with particularly dominant performance on heavily temporal benchmarks like TempCompass (75.1). This indicates that closely aligning the scoring mechanism with the policy's internal representations provides optimal guidance. However, using a 7B-parameter model introduces computational overhead. To address this, we evaluate lightweight scorers. Relying on SigLIP-2 provides efficient visual-semantic alignment but truncates lengthy video QA queries due to its limited context window (57.4). LLaVA-OneVision-0.5B (LLaVA-OV-0.5B) resolves this by striking an ideal balance: it is small enough for efficient scoring yet possesses a sufficiently large context window to process long queries in their entirety, recovering significant performance to achieve a competitive average of 58.0.

\begin{table}[t!]
    \centering
    \caption{\textbf{Ablation Studies.} We evaluate (a) the variant construction strategy, (b) the scoring mechanism used for importance-based grouping, (c) the trade-off between visual variants ($M$) and textual generations ($G$), and (d) the number of temporal segments ($K$). All ablation models are trained for 500 RL steps and evaluated on a subset of four core benchmarks.}
    \label{tab:combined_ablations}
    \resizebox{1.0\linewidth}{!}{%
    \begin{tabular}{l|c|c|c|c|c}
        \toprule
        Configuration & VideoMME & VSI-Bench & TempCompass & MMVU (mc) & Avg \\ 
        \midrule
        \multicolumn{6}{l}{\textit{(a) Variant Construction Strategies}} \\
        \midrule
        GRPO Baseline                              & 58.0              & 32.5           & 72.1    & 65.3      & 57.0 \\ 
        Deterministic Temporal                     & 57.1              & 31.8           & 73.1    & 65.0      & 56.8 \\ 
        Stochastic                                 & 58.4              & 34.3           & \textbf{73.2}    & 65.8      & 57.9 \\ 
        Importance-Based (Temporal Only)           & \textbf{59.0}     & 34.2           & 72.8    & \textbf{66.2}      & \textbf{58.1} \\ 
        Importance-Based (+ Spatial)               & 58.2              & \textbf{35.2}  & 73.1    & 65.4      & 58.0 \\
        \midrule
        \multicolumn{6}{l}{\textit{(b) Scoring Mechanism}} \\
        \midrule
        Policy Model (Grad)               & 61.1              & 32.2           & 72.8    & 64.8      & 57.7 \\
        Policy Model (Sim)                & \textbf{61.8}    & 33.4           & \textbf{75.1}    & 64.0      & \textbf{58.6} \\ 
        SigLIP-2 (Sim)                    & 60.7             & 32.1           & 72.4    & 64.4     & 57.4 \\ 
        LLaVA-OV-0.5B (Sim)               & 58.2              & \textbf{35.2}  & 73.1    & \textbf{65.4}      & 58.0 \\ 
        \midrule
        \multicolumn{6}{l}{\textit{(c) Visual Variants ($M$) vs. Textual Generations ($G$)}} \\
        \midrule
        $M=1, G=8$ (Baseline)       & 58.0     & 32.5      & 72.1        & 65.3      & 57.0 \\
        $M=2, G=4$ (STRIVE)         & \textbf{58.2} & \textbf{35.2} & \textbf{73.1} & \textbf{65.4} & \textbf{58.0} \\     
        $M=4, G=2$                  & 58.1 & 34.5 & 73.0  & 65.3 & 57.7 \\ 
        \midrule
        \multicolumn{6}{l}{\textit{(d) Number of Temporal Segments ($K$)}} \\
        \midrule
        $K=4$                       & 57.9          & 34.8          & 72.1          & 64.8          & 57.4 \\
        $K=8$ (STRIVE)              & \textbf{58.2} & \textbf{35.2} & 73.1          & \textbf{65.4} & \textbf{58.0} \\
        $K=16$                      & 58.1          & 34.3          & \textbf{73.5} & \textbf{65.4} & 57.8 \\
        \bottomrule
    \end{tabular}
    }
\end{table}

\bparagraph{Impact of Textual Generations vs. Visual Variants.}
We conduct an ablation study to determine the optimal configuration of visual variant count ($M$) and textual generation count ($G$) in our STRIVE framework (Table~\ref{tab:combined_ablations}c). To ensure a strictly fair comparison, we maintain a constant total computational budget of $M \times G = 8$ forward passes per training step. The baseline standard GRPO configuration, which uses a single visual representation ($M=1$) to sample $G=8$ textual generations, achieves an average performance of 57.0. By splitting the budget to $M=2$ distinct visual variants with $G=4$ generations each, STRIVE yields the best overall performance (58.0), with an improvement on spatial reasoning tasks like VSI-Bench (+2.7). However, aggressively partitioning the budget to $M=4$ visual variants with only $G=2$ generations per variant results in a slight performance drop to 57.7. This reveals an important dynamic: while visual diversity ($M$) prevents advantage collapse, the algorithm still requires a sufficient number of textual samples ($G$) within each group to compute a reliable advantage baseline. A balanced configuration provides the optimal trade-off.

\begin{figure}[t!]
   \centering
        \includegraphics[width=0.9\columnwidth]{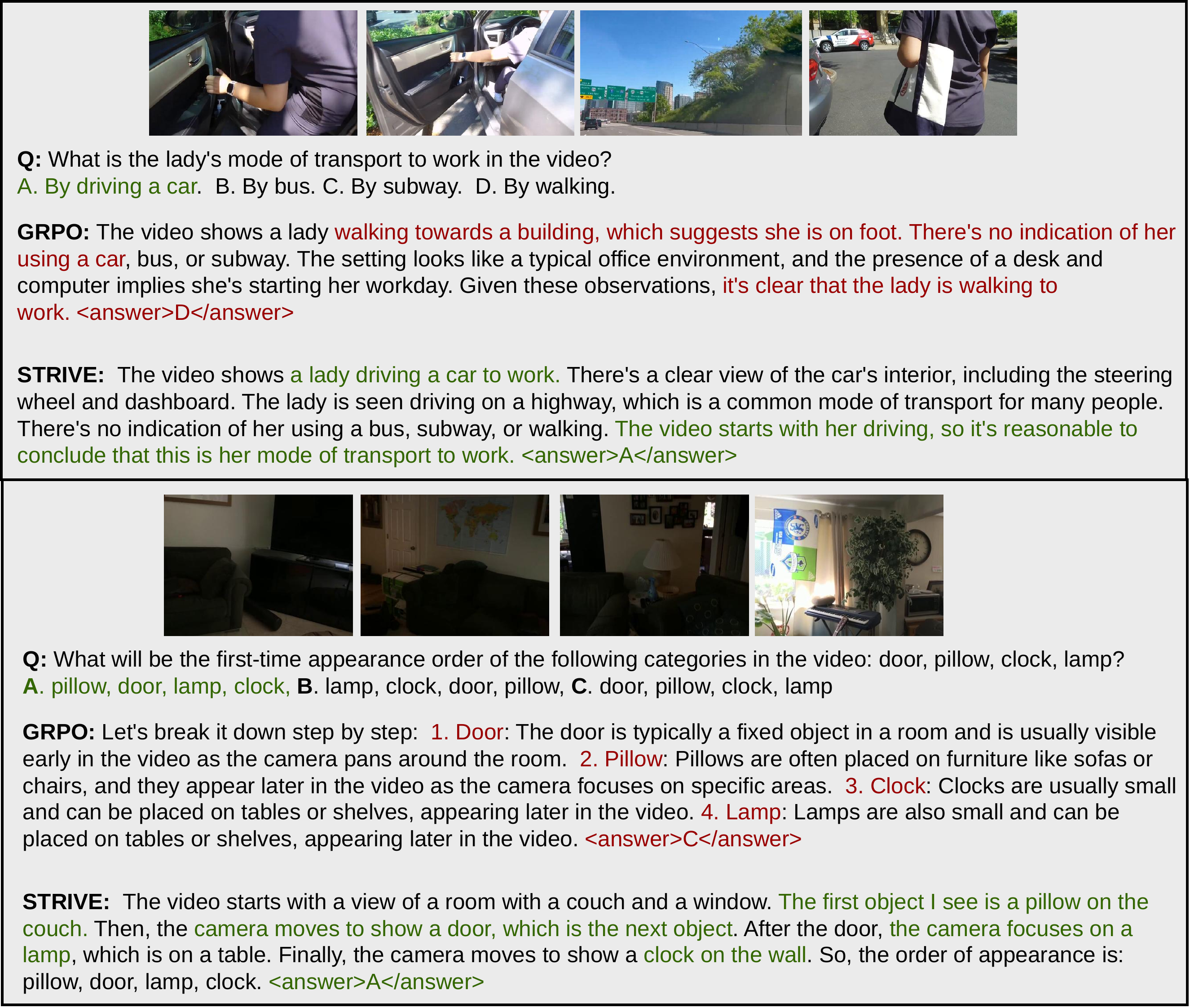}
        \caption{\textbf{Qualitative comparison between STRIVE and the standard GRPO baseline.} (Top) STRIVE successfully isolates the relevant driving frames to correctly answer a question about a person's mode of transport, whereas GRPO is distracted by irrelevant walking scenes. (Bottom) STRIVE accurately determines the strict temporal sequence of objects, demonstrating how question-guided variant construction improves fine-grained spatiotemporal reasoning and prevents reliance on flawed priors.}
        \label{fig:qual_result}
\end{figure}

\bparagraph{Impact of Temporal Segments ($K$).}
We evaluate the sensitivity of our importance-aware sampling to the number of temporal segments, $K$ (Table~\ref{tab:combined_ablations}d). To ensure a strictly fair comparison, we maintain a fixed budget of 16 total sampled frames across all configurations by adjusting the number of frames sampled per segment. We observe that $K=4$ (sampling 4 frames per segment) provides insufficient temporal granularity, leading to the lowest overall average (57.4) and struggling notably on temporal reasoning tasks like TempCompass (72.1). Conversely, an extremely fine-grained partition of $K=16$ (sampling exactly 1 frame per segment) excels on TempCompass (73.5) but overly restricts the model's intra-segment exploration, causing a performance drop on spatial tasks like VSI-Bench (34.3). Configuring $K=8$ achieves the optimal balance by sampling 2 frames per segment. This configuration preserves both structural temporal coverage and sufficient local flexibility to isolate critical visual evidence, yielding the highest overall average performance (58.0).

\bparagraph{Qualitative Analysis.}
Figure~\ref{fig:qual_result} presents qualitative examples highlighting the improved spatiotemporal reasoning of STRIVE compared to the standard GRPO baseline. In the first example (top), the model must determine the woman's mode of commute. STRIVE correctly identifies that she is driving by isolating the critical temporal window showing the car's interior and the highway. In contrast, the baseline GRPO model is distracted by irrelevant frames of her walking at other points in the video. In the second example (bottom), the task requires fine-grained temporal ordering. While standard GRPO fails by seemingly relying on a rigid, static object prior, STRIVE accurately tracks the strict temporal progression. By utilizing importance-based variant construction, STRIVE leverages question-specific semantic information to guide frame sampling, effectively filtering out temporal noise.

\section{Conclusion}
We presented STRIVE, a structured reinforcement learning framework for video question answering that integrates spatiotemporal variant construction into group-based policy optimization. By jointly normalizing textual generations and complementary visual variants, STRIVE enriches reward signals and stabilizes advantage estimation. An importance-aware sampling strategy further promotes robust multimodal reasoning by prioritizing query-relevant frames while preserving temporal coverage. Experiments across six challenging video benchmarks demonstrate consistent improvements over strong reinforcement learning baselines, highlighting structured spatiotemporal exploration as an effective mechanism for stabilizing multimodal policy optimization. 


\bibliographystyle{splncs04}
\bibliography{main}
\end{document}